# Improving Mortality Prediction After Radiotherapy with Large Language Model Structuring of Large-Scale Unstructured Electronic Health Records

Sangjoon Park, MD, PhD[1,2]; Chan Woo Wee, MD, PhD[1]; Seo Hee Choi, MD[1]; Kyung Hwan Kim, MD, PhD[1]; Jee Suk Chang, MD, PhD[1]; Hong In Yoon, MD, PhD[1]; Ik Jae Lee, MD, PhD[1]; Yong Bae Kim, MD, PhD[1]; Jaeho Cho, MD, PhD[1]; Ki Chang Keum, MD, PhD[1]; Chang Geol Lee, MD, PhD[1]; Hwa Kyung Byun, MD, PhD[3*]; Woong Sub Koom, MD, PhD[1*]

[1]Department of Radiation Oncology, Yonsei University College of Medicine, Seoul, Korea

[2]Institute for Innovation in Digital Healthcare, Yonsei University, Seoul, Korea

[3]Department of Radiation Oncology, Yongin Severance Hospital, Yongin, Gyeonggi-do, Korea

***Corresponding Authors**

Hwa Kyung Byun, MD, PhD, Department of Radiation Oncology, Yongin Severance Hospital, Email: HKBYUN05@yuhs.ac

Woong Sub Koom, MD, PhD, Department of Radiation Oncology, Yonsei University College of Medicine, Email: mdgold@yuhs.ac

**Abstract**

Accurate survival prediction in radiotherapy (RT) is critical for optimizing treatment decisions. This study developed and validated the RT-Surv framework, which integrates general-domain, open-source large language models (LLMs) to structure unstructured electronic health records alongside structured clinical data. Using data from 34,276 patients and an external cohort of 852, the framework successfully transformed unstructured clinical information into structured formats. Incorporating LLM-structured clinical features improved the concordance index from 0.779 to 0.842 during external validation, demonstrating a significant performance enhancement. Key LLM-structured features, such as disease extent, general condition, and RT purpose, showed high predictive importance and aligned closely with statistically significant predictors identified through conventional statistical analyses, thereby improving model interpretability. Furthermore, the framework enhanced risk stratification, enabling more distinct differentiation among low-, intermediate-, and high-risk groups ($p < 0.001$) using LLM-structured clinical features. These findings highlight the potential of LLMs to convert unstructured data into actionable insights, improving predictive modeling and patient outcomes in clinics.

## Introduction

Radiotherapy (RT) is an essential component in cancer treatment, with approximately 60% of cancer patients undergoing this procedure during their treatment course, according to the 2023 Radiation Oncology Case Rate Report[1]. Projections from the Surveillance, Epidemiology, and End Results database indicate that the number of patients receiving RT will rise to 3.38 million by 2020 and 4.17 million by 2030[2]. While the benefits of RT (e.g., symptom relief, improved survival) are well documented, they are influenced by factors such as tumor type, treatment site, and patient health status. Some patients may not live long enough to benefit from RT, making accurate selection of suitable patients crucial to avoid unnecessary treatments, burdens, and healthcare costs[3,4].

Several studies have predicted survival outcomes in patients receiving RT by focusing on short-term mortality factors or developing prognostic nomograms[5-7]. However, these methods often fail to accurately predict survival duration, thus limiting their practical utility in clinical decision-making. The advent of machine learning has enabled the exploration of survival prediction in patients receiving RT using electronic health record (EHR) data, which includes primarily structured data (e.g., patient demographics, vital signs, laboratory results)[8]. However, this approach neglects critical information found in unstructured clinical notes, such as disease extent, treatment purpose, and patient condition. Manual structuring of such unstructured data is impractical on a large scale.

Large language models (LLMs) such as OpenAI's ChatGPT have demonstrated significant capabilities in processing unstructured texts. These models can perform new tasks with few-shot learning, enabling data structuring without explicit training, heralding a new era of generative artificial intelligence models[9-11]. Their flexibility and adaptability, especially when well-structured prompts are used, make them ideal for structuring clinical records.

Consequently, there is growing interest in using LLMs for data standardization in the medical domain[12-14].

The current study aimed to develop a model that could predict post-RT mortality by leveraging comprehensive structured and unstructured data from patient records at a large-volume center. Using an open-source LLM that can be deployed with internal hospital resources, we ensured data privacy without risking the leakage of patient information. By structuring unstructured clinical data, we aimed to enhance survival prediction accuracy and provide guidelines on effective LLM utilization in clinical practice through data standardization, ultimately advancing patient outcomes. To validate the robustness of our findings, we also applied our model to an external database from another institution, demonstrating its generalizability and practical utility in diverse clinical settings.

## Methods

### Study design and participants

We developed a model to assist in clinical decision-making by providing survival estimates in RT consultation. Data were collected from patients treated at Yonsei Cancer Center between August 2013 and July 2023. The exclusion criteria were as follows: (1) incomplete radiation oncology records and (2) unconfirmed post-RT survival through the national insurance system. Out of 51,821 patients, 34,276 were used for LLM data structuring, whereas 25,183 were used for survival prediction, with 20% randomly reserved for internal validation (**Supplementary Fig. 1A**).

For external validation, data from 852 eligible patients treated at Yongin Severance Hospital between March 2020 and July 2023 were included (**Supplementary Fig. 1B**). Structured data (age, height, weight, BMI, vital signs, blood counts, and routine blood chemistry) and unstructured text data (medical records and imaging reports) were collected, prioritizing universally available tests.

This study was conducted in accordance with the principles embodied in the Declaration of Helsinki and approved by the Institutional Review Boards (IRBs) of Severance and Yongin Severance Hospitals (IRB approval numbers 4-2024-0671 and 9-2024-0122). The need for informed consent was waived. Further details are provided in **Supplementary Methods**.

**Data collection**

Using the Severance Clinical Research Analysis Portal (SCRAP), we automatically extracted structured and unstructured data from EHRs of the Yonsei University Healthcare System. These data included patient demographics and clinical records in RT consultation at the Department of Radiation Oncology, including referral reasons, medical history, clinical summaries, and treatment plans. Additionally, unstructured text reports from imaging studies closest to the consultation, such as PET-CT, chest CT, abdominopelvic CT, MRI, and chest and abdominal radiography, were obtained. Vital signs, physical measurements, complete blood counts, and routine blood chemistry results from the nearest outpatient visit were also recorded. Further details are provided in **Supplementary Methods**, **Supplementary Fig. 2**, and **Supplementary Table 1**.

**RT-Surv framework**

The proposed RT-Surv framework, compared with conventional methods, is shown in **Fig. 1**. While structured data are easily used for predictive modeling (**Fig. 1A**), unstructured text data in both English and Korean pose challenges. To address this, we incorporated open-source pretrained LLMs into the RT-Surv framework to structure EHR data (**Fig. 1B**).

While proprietary LLMs such as GPT-4, Gemini 1.5 Pro, and Claude 3.5 Sonnet offer superior performance, they raise privacy concerns owing to external data transmission[15]. To mitigate this, we utilized open-source LLMs on institutional resources, specifically Meta's LLaMA-3 model (8B and 70B), and compared them with a domain-specific model (Med-LLaMA)[16].

Using a single-shot learning approach with expert-designed prompts, the LLM categorized EHR data into general condition, tumor pathology, disease extent, control trend, RT purpose, prior RT history, and urgency. This process relied on radiation oncology records, including referral reasons, medical history, treatment plans, and imaging reports.

Finally, we developed a predictive model combining structured EHR data with LLM-extracted features from unstructured data and compared its performance to that of a model based solely on structured data to evaluate the benefits of LLM-driven data structuring. Details on framework development, implementation, and structuration are provided in **Supplementary Methods** and **Supplementary Table 2**.

**Prediction models and benchmarking**

Within the RT-Surv framework, we developed and benchmarked predictive models using three approaches: the Cox proportional hazards (Cox PH) model, which represents a statistical method; the random survival forest (RSF) model, which is based on machine learning; and the DeepSurv model, which utilizes deep learning.[17] This comprehensive evaluation sought to determine the extent to which the inclusion of unstructured EHR data structured through the application of LLM enhances model performance across these varied analytical approaches.

To prevent performance degradation and overfitting due to missing data, we first identified the features associated with short-term mortality. Guided by the UK's National Health Service 30-day mortality (30-DM) rate, we initially selected features showing significant differences in relation to 30-DM occurrence.[18] We further refined the selection by assessing the correlation between each feature and 30-DM using Kendall's Tau rank correlation, including only features with an absolute correlation value of 0.1 or greater in the modeling analysis.

**Risk group stratification**

To evaluate the concordance between survival predictions generated by the prediction model and actual patient outcomes, a risk group-based analysis was conducted. The prediction model provided direct survival probabilities at specified time points, enabling risk group classification based on predicted probabilities. Risk groups were defined as follows based on the predicted survival probability at 90 days: low-risk (≥80%), intermediate-risk (60–80%), and high-risk (<60%). Patients were stratified into these groups, and differences in actual survival outcomes among the risk groups were analyzed to assess the validity of the model's risk predictions.

**Evaluation of LLM Accuracy in single-shot structuration**

To assess the accuracy of the LLMs in single-shot structuration, a board-certified radiation oncologist selected 20 patient cases from the entire dataset, encompassing a range of RT scenarios and patient conditions. Patients with insufficient unstructured data for accurate structuring were excluded.

The accuracy of the LLM-structured clinical features was evaluated across the aforementioned seven categories, with each category assessed on a binary scale (0 = incorrect, 1 = correct). Two board-certified radiation oncologists, each with over 5 years' experience and from different centers, conducted the evaluations independently. The evaluators were blinded to each other's assessments to ensure an unbiased and rigorous evaluation of the LLM-structured clinical features.

**Statistical analysis**

To assess the accuracy of the LLM-generated summaries, we calculated the accuracy of each of the seven categories. Differences in features based on 30-DM occurrence were visualized using box plots for continuous variables and stacked bar plots for categorical variables. The survival prediction accuracy was evaluated using three primary metrics: Harrell's concordance index (C-index), integrated Brier score (IBS), and negative binomial log-likelihood (NBLL). Survival were estimated using the Kaplan-Meier method, and differences between groups were compared using the log-rank test. The Cox proportional hazards model was employed to calculate the log hazard ratios and identify statistically significant prognostic factors for survival. Confidence intervals (CIs) for each metric were calculated using a nonparametric bootstrap method with 1,000 random samples drawn with replacement. The mean values and 95th percentile CIs were estimated from the relative frequency distribution of each trial. Non-overlapping CIs or $P$ values <.05 were considered significant.

## Results

**Patient characteristics and clinical correlates of 30-day mortality**

The demographics of 34,276 patients derived from directly extracted structured EHR data are presented in **Supplementary Table 3**. Among patients who developed 30-DM post-RT, overall poorer characteristics were lower body mass index, faster pulse rate, lower blood cell counts, higher inflammation markers, elevated liver enzymes, and lower levels of plasma protein and albumin. Specifically, these patients exhibited a higher pulse rate (Kendall's tau correlation coefficient 0.095), absolute neutrophil count (0.112), neutrophil-lymphocyte ratio (0.156), and alkaline phosphatase level (0.129). Conversely, they showed lower levels of red blood cells (RBC) (-0.101), hemoglobin (-0.105), hematocrit (-0.106), total protein (-0.102), absolute lymphocyte count (-0.110, *P*<.001), albumin (-0.169), sodium (-0.154), and chloride (-0.117) (all *P*<.001) (**Supplementary Fig. 3**).

**Accuracy of LLMs in structuring unstructured EHR data**

**Table 1** shows the accuracy of the LLMs in structuring unstructured EHR data across seven LLM-structured clinical features as evaluated by clinical experts. General-purpose, non-fine-tuned LLMs such as LLaMA-3-70B exhibited significantly higher accuracy than domain-specific, fine-tuned models such as Med-LLaMA, which struggled with more complex prompts (**Supplementary Table 4**). The average accuracy across seven clinical features was 87.5% (95% CI, 83.6–91.1) for LLaMA-3-70B, whereas Med-LLaMA achieved only 35.8% (95% CI, 30.4–41.4). Models with larger parameters demonstrated superior accuracy compared with smaller models, with LLaMA-3-70B and LLaMA-3-8B achieving 87.5% (95% CI, 83.6–91.1) and 70.7% (95% CI, 65.4–76.1), respectively. The highest accuracy was noted in classifying primary pathology (100.0%; 95% CI, 100.0–100.0), with high accuracies also observed in determining the disease extent (92.5%; 95% CI, 82.5–100.0), disease control trends (94.9%; 95% CI, 87.5–100.0), and aim of RT (92.6%; 95% CI, 84.9–100.0). However, lower accuracy was seen in determining the urgency of RT (84.8%; 95% CI, 72.5–95.0) and identifying whether the current RT session was a re-irradiation (82.4%; 95% CI, 70.0–92.5). The lowest accuracy was observed in assessing the general condition of patients, with an average accuracy of 65.2% (95% CI, 50.0–80.0).

**Analyses and implications of LLM-structured clinical features**

Using LLM-structured data, we identified key characteristics of patients who developed 30-DM post-RT. Demographics, including variables structured by the LLM from unstructured EHR data, are presented in **Supplementary Table 5**. The LLM identified that patients who developed 30-DM post-RT generally had poorer general condition, more extensive disease, poorer disease control, higher palliative RT rates, higher re-irradiation rates, and more urgent RT needs. Specifically, these patients were more likely to have a poorer general condition (Kendall's tau correlation coefficient 0.145), more extensive disease (0.191), poor disease control (0.063), receive palliative rather than curative RT (0.221), undergo re-irradiation (0.062), and require more urgent treatment (0.175) (all $P$<.001) (**Fig. 2**).

**Performance improvements with RT-Surv framework in survival prediction models**

Incorporating LLM-structured clinical features into predictive models led to substantial performance improvements across both internal and external validation cohorts (**Table 2** and **Supplementary Fig. 4**). During internal validation, the C-index for the Cox PH model was increased from 0.710 (95% CI, 0.699–0.719) to 0.809 (95% CI, 0.801–0.817), IBS improved from 0.196 (95% CI, 0.192–0.201) to 0.136 (95% CI, 0.130–0.142), and NBLL decreased from 0.570 (95% CI, 0.558–0.582) to 0.422 (95% CI, 0.405–0.440). Similarly, the RSF model demonstrated an increase in C-index from 0.710 (95% CI, 0.700–0.719) to 0.809 (95% CI, 0.801–0.817), alongside a reduction in IBS from 0.196 (95% CI, 0.191–0.201) to 0.136 (95% CI, 0.130–0.142) and a decrease in NBLL from 0.570 (95% CI, 0.558–0.582) to 0.422 (95% CI, 0.406–0.439). The DeepSurv model also exhibited significant enhancements, with the C-index increasing from 0.737 (95% CI, 0.727–0.746) to 0.820 (95% CI, 0.813–0.827),

IBS improving from 0.183 (95% CI, 0.177–0.190) to 0.131 (95% CI, 0.125–0.137), and NBLL decreasing from 0.546 (95% CI, 0.527–0.566) to 0.409 (95% CI, 0.391–0.427). These improvements were corroborated by external validation using data from another institution, where the C-index for the Cox PH model increased from 0.744 (95% CI, 0.717–0.769) to 0.837 (95% CI, 0.820–0.854). The RSF model also showed enhancement, with the C-index increasing from 0.744 (95% CI, 0.719–0.768) to 0.837 (95% CI, 0.819–0.855), and the DeepSurv model demonstrated an increase from 0.779 (95% CI, 0.755–0.800) to 0.842 (95% CI, 0.823–0.859). Additionally, IBS values were improved across all models during external validation, with the Cox PH and the RSF models both improving from 0.175 (95% CI, 0.164–0.188) to 0.130 (95% CI, 0.116–0.143), and the DeepSurv model from 0.164 (95% CI, 0.148–0.182) to 0.128 (95% CI, 0.112–0.144). The NBLL also demonstrated improvement across all models, decreasing from 0.526 (95% CI, 0.498–0.558) to 0.399 (95% CI, 0.364–0.433) in the Cox PH model, from 0.526 (95% CI, 0.497–0.557) to 0.398 (95% CI, 0.363–0.435) in the RSF model, and from 0.507 (95% CI, 0.461–0.554) to 0.399 (95% CI, 0.354–0.443) in the DeepSurv model.

**Enhanced risk stratification with RT-Surv framework and unstructured EHR data**

To evaluate the predictive performance of the proposed RT-Surv model, survival outcomes were analyzed for both internal and external validation cohorts, stratified by predicted risk groups. **Fig. 3** presents Kaplan-Meier survival curves for the internal and external validation sets, comparing the use of structured data alone with the combination of structured and unstructured data. Survival curves for the internal validation set using only structured EHR data (**Fig. 3A**) show that the model significantly distinguished between low, intermediate, and high-risk groups (all $P < 0.001$). However, in external validation (**Fig. 3B**), while the low-risk group remained distinct (all P < 0.001), separation between intermediate and high-risk groups was not statistically significant (P = 0.115). These results indicate that the structured-only model has limitations in capturing comprehensive patient characteristics, especially in external validation.

Survival curves for models combining structured EHR data with LLM-structured unstructured data are shown for internal (**Fig. 3C**) and external (**Fig. 3D**) validation sets. The inclusion of unstructured data significantly improved model performance, resulting in clearer stratification among risk groups. In both cases, survival curves for low, intermediate, and high-risk groups were well-separated, with statistically significant differences (all P < 0.001) in both internal and external validations. This enhancement demonstrates the model's improved ability to identify patients with varying prognoses, particularly those at the highest risk. These findings highlight that integrating LLM-structured unstructured data into the RT-Surv framework enhances the model's risk assessment capabilities, enabling more accurate patient stratification and better-informed treatment planning.

**Feature importance demonstrates consistency with statistical factors for survival**

In the feature importance analysis of the prediction models (**Fig. 4**), we examined the relationship between statistically significant survival factors and their corresponding feature importance. The analysis demonstrated that, when using only structured data, factors such as albumin, RBC, and total protein were statistically significant for survival (**Fig. 4A**) and also showed high feature importance in the prediction models (**Fig. 4B-D**). Similarly, when both structured and unstructured data were used, factors such as general condition, aim of RT, emergency status, RBC, and albumin remained statistically significant for survival (**Fig. 4E**) and also exhibited high feature importance across all models (**Fig. 4F-H**). Incorporating

LLM-structured clinical features emphasized these key variables, making them more influential than traditional laboratory markers like albumin, RBC, and alkaline phosphatase in predicting post-RT survival. The addition of unstructured data increased the feature importance of clinically meaningful factors such as general condition, disease extent, and aim of RT. The alignment between statistical significant factors and feature importance indicates that the models effectively reference clinically relevant factors, enhancing their interpretability and providing greater confidence in their application for clinical decision-making.

**Discussion**

This study represents a significant advancement in the application of LLMs in medicine. Previous studies have predominantly focused on evaluating how well LLMs encode clinical knowledge, often employing them for tasks such as question answering or summarization[19-24]. Although valuable for assessment purposes, these applications have limited practical utility in clinical settings and typically do not enhance patient outcomes or significantly improve clinical practice. However, our study is the first to demonstrate that the appropriate application of LLMs can improve patient prognosis and the quality of healthcare delivery. Specifically, the integration of unstructured data through the RT-Surv framework enhanced predictive accuracy, particularly by enabling more precise risk group stratification. This enhanced risk group stratification with RT-Surv framework provided clearer distinctions in patient survival and was especially pronounced in external validation cohorts, highlighting the model's strong generalizability. This study offers a novel perspective on the effective utilization of LLMs to improve clinical outcomes and provides a guide for future applications across diverse healthcare settings.

Our findings suggest that LLMs have the potential to process extensive unstructured data, which would be impractical for manual structuring, thereby advancing the development of sophisticated predictive models. Notably, general-domain, open-source LLMs demonstrated high accuracy in structuring unstructured data, even without extensive tuning, using a single-shot example approach. This capability was evident in the clinically predictable trends observed in the structured data. For instance, higher 30-DM rates were noted among patients on palliative care, those in poorer general condition, and those with extensive disease. These trends align with clinical expectations, reinforcing the potential of LLMs to transform unstructured EHR data into a structured format that is both reliable and actionable for clinical decision-making.

Despite these promising results, this study also highlighted areas in which LLMs have limitations. While LLMs performed well in distinguishing clear-cut clinical variables, such as primary pathology, they struggled with more complex tasks requiring comprehensive contextual understanding, such as integrating longitudinal data and anatomical correlations. For example, there were significant challenges in identifying previous RT fields that require an understanding of anatomical details combined with historical treatment records. Additionally, the accuracy in assessing the general condition was the lowest, reflecting the difficulty in accurately inferring a patient's condition from imaging results or clinical records without direct clinical interviews. These limitations highlight the difficulties faced by LLMs in handling tasks that require extensive anatomical knowledge or complex contextual understanding, or ambiguous tasks owing to insufficient information. Nonetheless, reasonable trends were captured by the LLMs, which contributed to the substantially enhanced performance of the predictive models.

In addition to improving the predictive model performance, the inclusion of LLM-structured clinical features not only improved the predictive performance of the models but also significantly enhanced their interpretability. Clinical factors such as the patient's general condition, the aim of RT, and emergency indicators, when structured by the LLM, demonstrated strong statistical correlations with patient survival. These factors were also highlighted as highly important in the model's feature importance analysis, exhibiting a robust association with the significant prognostic factors in statistical analysis. In addition to traditionally structured EHR data variables, such as albumin and RBC counts, which are established survival factors, the integration of previously unstructured clinical features further aligned with known clinical expectations and known relevance to patient survival. This improvement in interpretability suggests that structured data derived from unstructured text enhance predictive accuracy and provide more clinically meaningful insights by quantifying the relative importance of clinically significant factors not originally structured in EHR data[25,26].

Our findings challenge the prevailing assumption that domain-specific models, including medically fine-tuned LLMs, are inherently superior for processing specialized domain data[27-29]. Contrary to this belief, we observed that general-domain LLMs, even without fine-tuning for the medical domain, performed exceptionally well. This contrasts with the approach adopted in many studies that emphasized tuning LLMs for specific fields, such as medicine, which, while potentially enhancing domain-specific knowledge, often diminishes language adaptability[30-32]. For instance, Med-LLaMA, a medically fine-tuned model, struggled to process complex prompts effectively, as shown in **Supplementary Table 4**. These results suggest that open-source LLMs optimized for general language comprehension and prompt adherence may be more effective and clinically valuable by employing single-shot learning than the conventional approach of extensive domain-specific tuning.

The framework developed in this study, RT-Surv, has broad applications beyond the field of radiation oncology. Unstructured clinical records are the fundamental form of EHR data across all medical specialties and are not limited to radiation oncology[33]. Therefore, this framework can be adapted for various medical fields to reduce overall hospital mortality rates and in scenarios requiring the integration and analysis of large-scale clinical data. The ability to automatically structure vast amounts of unstructured data facilitates more accurate and efficient predictive modeling across healthcare settings.

This study had some limitations. First, the accuracy of data structured by the LLMs was approximately 90%. Although the overall trends were interpretable and reliable, they might not have achieved optimal performance compared with manually entered data. Second, there was a significant amount of missing data. While tree-based models such as RSF managed this effectively, models such as Cox PH and DeepSurv had to impute missing values with mean values, which may introduce biases and affect model robustness. Finally, while we validated our findings using an external dataset from another institution, further external validation across more diverse clinical settings is necessary to fully establish the generalizability of results.

In conclusion, this study demonstrated the effective integration of LLMs into predictive modeling, highlighting their potential to handle unstructured data and improve clinical outcomes. Our research shows that general-domain, open-source LLMs can accurately structure unstructured clinical data, leading to significantly enhanced performance in predictive models. These findings offer valuable insights into the future applications of LLMs in healthcare, extending beyond radiation oncology to benefit the broader medical field. This study highlights the utility of LLMs and their capacity to enhance healthcare quality in clinical practice.

**Author Contributions:** SJP contributed to concept development, data collection, analysis, and drafting and revision of the manuscript. HKB was involved in refining the research concept, data collection and analysis, and manuscript drafting and revision. WSK formulated the research concept, proposed the research questions, and revised the manuscript. SJP, HKB, and WSK approved the final version of this manuscript. All authors contributed to data collection, manuscript revision, and approval of the final manuscript, with full access to all data.

**Competing Interests:** We declare that there are no conflicts of interest.

**Acknowledgments:** This research was funded in part by the Basic Science Research Program through the National Research Foundation of Korea (NRF), funded by the Ministry of Education under grant number RS-2023-00242164; in part by the NRF grant funded by the Korean government (MSIT) under grant number RS-2024-00349635; and in part by a faculty research grant from the Yonsei University College of Medicine (6-2019-0153).

**Role of the Funder/Sponsor:** The funders had no role in the design and conduct of the study; collection, management, analysis, and interpretation of the data; preparation, review, or approval of the manuscript; and decision to submit the manuscript for publication.

**Data Availability:** The datasets used and/or analyzed during the current study available from the corresponding author on reasonable request.

**Code Availability:** The PyTorch codes for the proposed RT-Surv framework used in this study is available at the following GitHub repository: https://github.com/sangjoon-park/RT-Surv.

**Figure 1. Comparison of the conventional approach and the RT-Surv framework**

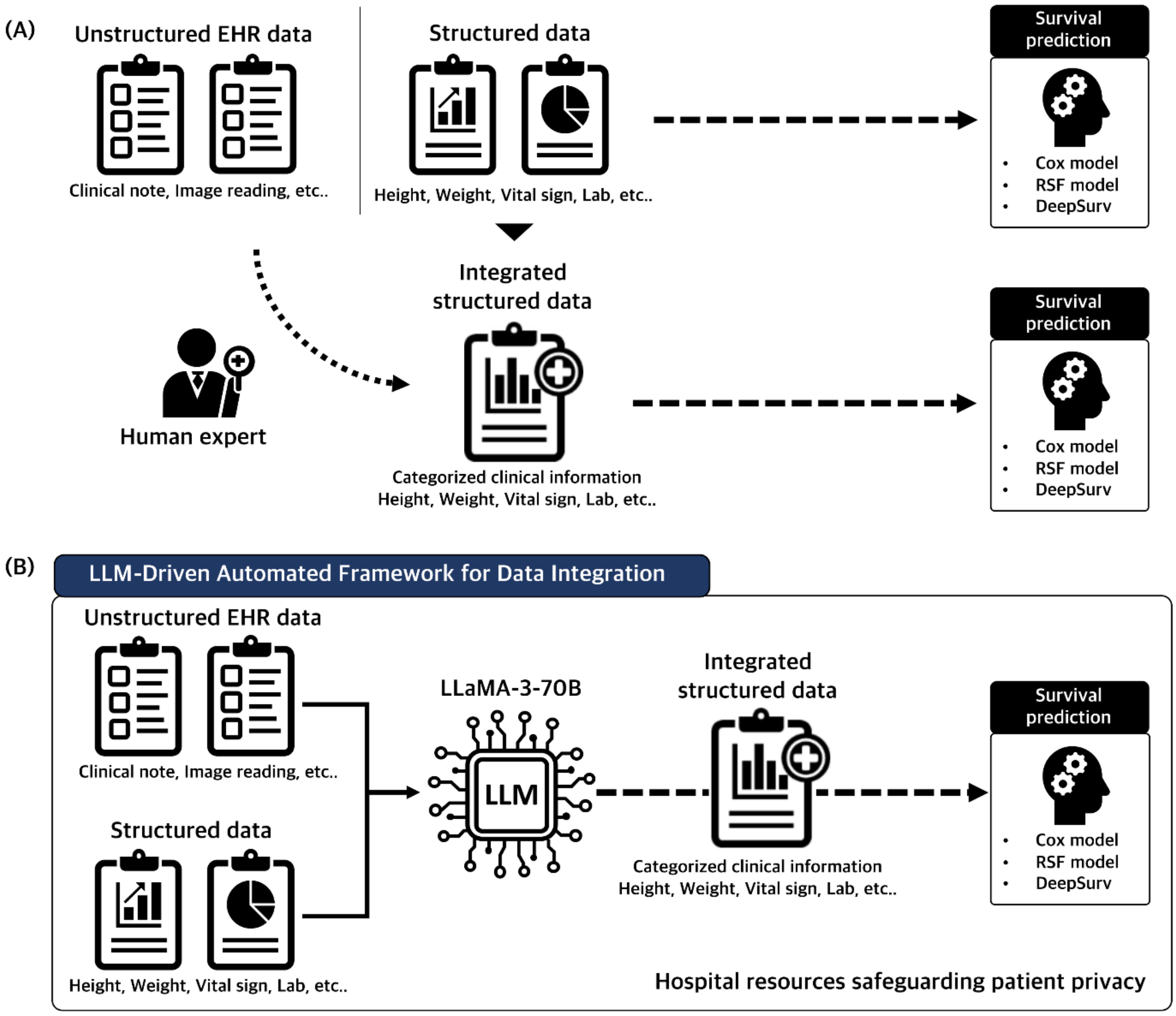

(A) The conventional approach uses only structured data or relies on human experts to structure unstructured data for model development. (B) The RT-Surv framework utilizes an open-source large language model (LLM) to automatically structure unstructured data and integrate them with structured data for predictive model development. RT, radiotherapy.

**Figure 2. Comparison of large language model (LLM)-structured clinical features from unstructured electronic health record (EHR) data according to 30-day mortality (30-DM) after radiotherapy (RT)**

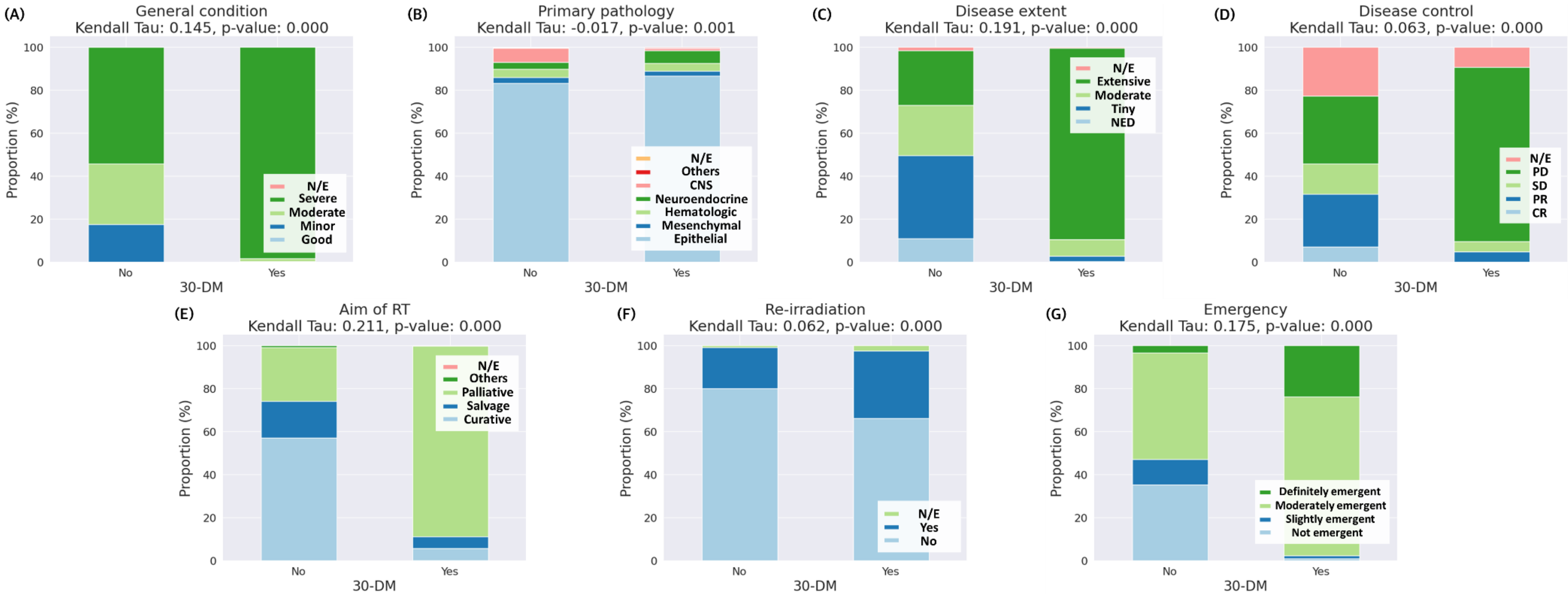

Comparison of clinically relevant categories, structured by an LLM from unstructured EHR data, stratified by the occurrence of 30-DM following RT. The bar plots represent the proportion of each class within the following categories: (A) general condition, (B) primary pathology, (C) disease extent, (D) disease control, (E) aim of RT, (F) re-irradiation, and (G) emergency of RT. Patients with 30-DM were more likely to have poor general condition, extensive disease, suboptimal disease control, palliative RT intent, and emergency RT. Kendall's Tau and p-values indicate the strength and significance of associations between clinical features and 30-DM. Abbreviations: N/E, not evaluable; CNS, central nervous system; NED, no evidence of disease; PD, progressive disease; SD, stable disease; PR, partial response; CR, complete response.

**Figure 3. Survival analysis comparing risk stratification by prediction models using structured electronic health record (EHR) data alone or with large language model (LLM)-structured clinical features from unstructured EHR data**

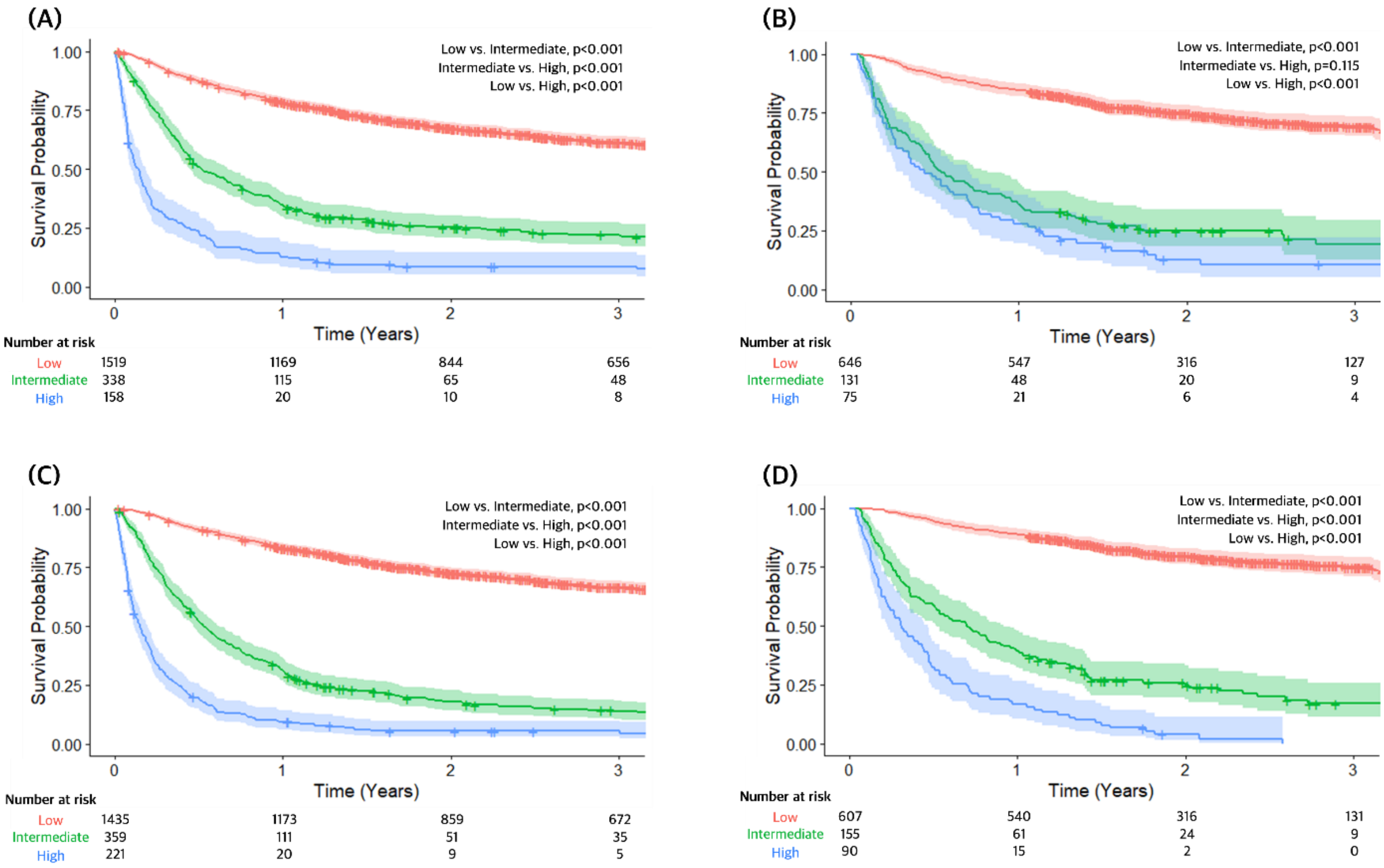

Kaplan-Meier survival curves stratified by predicted risk groups (low, intermediate, and high) for internal and external validation cohorts using structured EHR data alone or combined with LLM-structured unstructured data. (A) Internal validation set using structured EHR data alone shows significant separation among low, intermediate, and high-risk groups (all P < 0.001). (B) External validation set using structured data alone maintains significant separation for the low-risk group (all P < 0.001), but the distinction between intermediate and high-risk groups is not statistically significant (P = 0.115). (C) Internal validation using combined structured and LLM-structured unstructured data shows significant improvement, with clear stratification among all risk groups (all P < 0.001). (D) External validation using combined data similarly demonstrates significant separation among all risk groups (all P < 0.001).

**Figure 4. Feature importance of various prediction models using structured electronic health record (EHR) data alone or with large language model (LLM)-structured clinical features from unstructured EHR data**

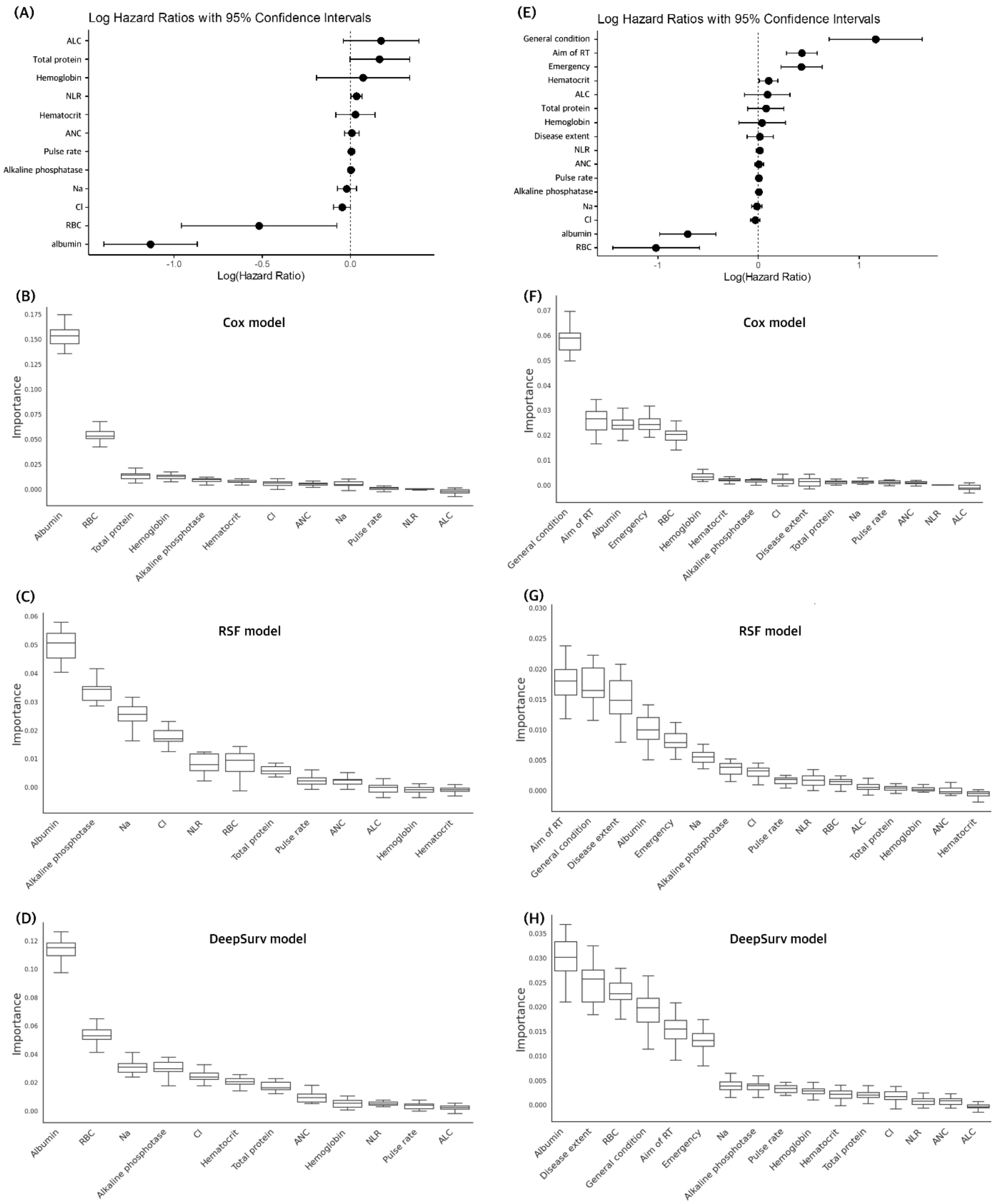

(A) Log hazard ratios for survival, derived from structured EHR data, analyzed using Cox proportional hazard regression. Box plots of feature importance in prediction models using only structured EHR data: (B) Cox model, (C) RSF model, and (D) DeepSurv model. (E) Log hazard ratios for survival, analyzed using both structured EHR data and LLM-structured clinical features from unstructured data, through Cox proportional hazard regression. Box plots of feature importance in prediction models incorporating both structured EHR data and LLM-structured clinical features: (F) Cox model, (G) RSF model, and (H) DeepSurv model. The results indicate that factors significantly influencing survival in statistical analysis were also identified as important features in prediction models, highlighting consistency between statistical significance and model-derived feature importance. RSF, random survival forest; RBC, red blood cell; Cl, chloride; ALC, absolute lymphocyte count; ANC, absolute neutrophil count; Na, sodium; NLR, neutrophil-to-lymphocyte ratio; RT, radiotherapy.

**Table 1. Accuracy of large language model (LLM) in structuring unstructured electronic health evaluated by clinical experts**

| Models | Accuracy (95% CI) | | | | | | | |
|---|---|---|---|---|---|---|---|---|
| | **General condition** | **Primary pathology** | **Disease extent** | **Disease control** | **Aim of RT** | **Re-irradiation** | **Emergency** | **Average** |
| LLaMA-3-70B | 65.2 (50.0-80.0) | 100.0 (100.0-100.0) | 92.5 (82.5-100.0) | 94.9 (87.5-100.0) | 92.6 (84.9-100.0) | 82.4 (70.0-92.5) | 84.8 (72.5-95.0) | 87.5 (83.6-91.1) |
| LLaMA-3-8B* | 70.1 (55.0-85.0) | 100.0 (100.0-100.0) | 75.3 (60.0-87.5) | 79.9 (65.0-92.5) | 70.1 (55.0-82.5) | 32.9 (20.0-47.5) | 67.5 (50.0-82.5) | 70.7 (65.4-76.1) |
| Med-LLaMA† | 39.9 (25.0-55.0) | 50.1 (32.5-67.5) | 25.3 (12.5-40.0) | 19.8 (7.5-32.5) | 29.6 (15.0-45.0) | 45.2 (30.0-60.0) | 40.0 (25.0-55.0) | 35.8 (30.4-41.4) |

CI, confidence interval; RT, radiotherapy; B, billion parameters. * This model is compared with a smaller model. †The model is also used for comparison with domain-specific model.

**Table 2. Performance comparison of prediction models using structured electronic health record (EHR) data alone or with large language model (LLM)-structured clinical features**

| Methods | Features | Cox PH model | RSF model | DeepSurv model |
|---|---|---|---|---|
| *Internal validation* | | | | |
| C-index* | Structured EHR data | 0.710 (0.699-0.719) | 0.710 (0.700-0.719) | 0.737 (0.727-0.746) |
| | Structured EHR data + LLM-structured clinical features | 0.809 (0.801-0.817) | 0.809 (0.801-0.817) | 0.820 (0.813-0.827) |
| IBS† | Structured EHR data | 0.196 (0.192-0.201) | 0.196 (0.191-0.201) | 0.183 (0.177-0.190) |
| | Structured EHR data + LLM-structured clinical features | 0.136 (0.130-0.142) | 0.136 (0.130-0.142) | 0.131 (0.125-0.137) |
| NBLL† | Structured EHR data | 0.570 (0.558-0.582) | 0.570 (0.558-0.582) | 0.546 (0.527-0.566) |
| | Structured EHR data + LLM-structured clinical features | 0.422 (0.405-0.440) | 0.422 (0.406-0.439) | 0.409 (0.391-0.427) |
| *External validation* | | | | |
| C-index* | Structured EHR data | 0.744 (0.717-0.769) | 0.744 (0.719-0.768) | 0.779 (0.755-0.800) |
| | Structured EHR data + LLM-structured clinical features | 0.837 (0.820-0.854) | 0.837 (0.819-0.855) | 0.842 (0.823-0.859) |
| IBS† | Structured EHR data | 0.175 (0.164-0.188) | 0.175 (0.163-0.188) | 0.164 (0.148-0.182) |
| | Structured EHR data + LLM-structured clinical features | 0.130 (0.116-0.143) | 0.130 (0.116-0.144) | 0.128 (0.112-0.144) |
| NBLL† | Structured EHR data | 0.526 (0.498-0.558) | 0.526 (0.497-0.557) | 0.507 (0.461-0.554) |
| | Structured EHR data + LLM-structured clinical features | 0.399 (0.364-0.433) | 0.398 (0.363-0.435) | 0.399 (0.354-0.443) |

PH, proportional hazards; RSF, random survival forest; C-index, concordance index; IBS, integrated Brier score; NBLL, negative binomial log-likelihood. * Higher values indicate better performance. †A lower value indicates better performance.

# Supplementary Material

**Supplementary Methods**

**Supplementary Table 1. Proportional of missing data for each electronic health record data features**

**Supplementary Table 2. Detailed prompts used for each clinical feature**

**Supplementary Table 3. Patient characteristics from structured electronic health record data according to 30-day mortality (DM) after radiotherapy (RT)**

**Supplementary Table 4. Example of input data, corresponding prompt, and responses from various large language models**

**Supplementary Table 5. Patient characteristics from structured electronic health record (EHR) data according to 30-day mortality (DM) after radiotherapy (RT)**

**Supplementary Fig. 1. Flow chart of patient selection and data split**

**Supplementary Fig. 2. Data collection timepoints of electronic health record (EHR) data**

**Supplementary Fig. 3. Comparison of structured electronic health record (EHR) data according to 30-day mortality (30-DM) after radiotherapy (RT)**

**Supplementary Fig. 4. Performance comparison of survival prediction models in internal and external validation**

## Supplementary Methods

### Details of study design and participants

We used data from a single large-volume center to create a model that could aid clinical decision-making by providing estimated survival predictions at the time of consultation for radiotherapy (RT), thereby helping practicing physicians make informed treatment decisions.

Data were collected from patients who underwent RT at the Yonsei Cancer Center between August 2013 and July 2023. Patients were excluded if (1) they had incomplete radiation oncology records that hindered the ability of large language model (LLM) to structure data or (2) their post-RT survival could not be confirmed through the national insurance system. Of the 51,821 patients treated, 34,276 were included in the LLM structuration analysis and 25,183 were included in the survival prediction analysis. Among the data, 20% was randomly reserved for internal validation, with no overlap of unique patient identifiers (**Supplementary Fig. 1A**).

For external validation, we included data from patients who received RT at Yongin Severance Hospital between March 2020 and July 2023. Of the 1,491 patients treated, 852 eligible patients were included in the external validation analysis for survival prediction (**Supplementary Fig. 1B**).

Structured data (age, height, weight, body mass index, vital signs, complete blood cell count, and routine blood chemistry results) and unstructured data (text-based medical records and imaging reports) were collected. To ensure broad applicability, we included only test results that were universally available across all patients, excluding cancer-specific tumor markers.

The study was conducted in accordance with the principles of the Declaration of Helsinki and was approved by the Institutional Review Boards (IRB) of Severance Hospital (IRB number 4-2024-0671) and Yongin Severance Hospital (IRB number 9-2024-0122). The requirement for informed consent was waived owing to the retrospective nature of the study.

### Details of data collection

Data were automatically extracted using the Severance Clinical Research Analysis Portal (SCRAP) from the Yonsei University Healthcare System. SCRAP is capable of extracting both structured and unstructured data from the electronic health record (EHR) system. Using SCRAP, we extracted basic patient information and clinical records from the day of RT consultation in the Department of Radiation Oncology, including referral reasons, medical history, clinical summaries, and treatment plans. Additionally, we extracted unstructured text reports of imaging studies (positron emission tomography-computed tomography [CT], chest CT, abdominopelvic CT, magnetic resonance imaging, and chest and abdominal radiographs) performed closest to the consultation date. Vital signs, physical measurements, complete blood cell counts, and routine blood chemistry results closest to the outpatient visit were also recorded.

The survival duration in patients was calculated from the date of RT initiation to that of death, as confirmed by the national insurance registration system.

To develop a model that aids in deciding whether to proceed with RT by predicting survival outcomes at the time of RT referral, we collected the closest patient data at the point of RT referral. This included anthropometric measurements, complete blood count (CBC), routine blood chemistry examination, and imaging test results.

For anthropometric measurements, we used the values obtained closest to the outpatient visit day, within a window ranging from 28 days before to 7 days after the visit. For CBC and routine blood chemistry examinations, which can change rapidly, we selected values from within 14 days before to 14 days after the outpatient visit, choosing the closest values to the visit day. For imaging tests, results within 28 days before to 7 days after the outpatient visit were included, with preference given to those closest to the visit day. A schematic representation of the data collection timepoints is provided in **Supplementary Fig. 2**.

Structured EHR data included height, weight, blood pressure (systolic and diastolic), and pulse rate for anthropometric measurements. CBC parameters collected were white blood cell (WBC) count, red blood cell (RBC) count, hemoglobin (Hb), hematocrit (Hct), platelet (PLT) count, absolute neutrophil count, absolute lymphocyte count, absolute monocyte count, absolute eosinophil count, and absolute basophil count. Routine blood chemistry examination results included plasma calcium, inorganic phosphate, glucose, blood urea nitrogen (BUN), creatinine, uric acid, cholesterol, total protein, albumin, alkaline phosphatase (ALP), aspartate aminotransferase (AST), alanine aminotransferase (ALT), total bilirubin, gamma-glutamyl transferase (GGT), sodium, potassium, chloride, prothrombin time (INR), and activated partial thromboplastin time (aPTT).

Unstructured EHR data collection involved imaging reports from brain magnetic resonance imaging, chest CT, abdominal-pelvic CT, and positron emission tomography-computed tomography to assess overall disease extent and control, as well as chest and abdominal radiographs to evaluate the patient's general condition. Additionally, clinical notes were collected, including detailed information on the patient's current disease status, reasons for referral, current diagnoses, disease progression, and the radiation oncologist's treatment plan. Not all data were available for every patient, and the proportion of missing data for each variable is provided in **Supplementary Table 1**.

**Details of RT-Surv framework development and implementation**

The RT-Surv framework proposed in this study, along with a comparison with conventional methods, is shown in **Fig. 1**. While structured data are readily applicable to predictive model development (**Fig. 1A**), unstructured text-based data, which encompass both English and Korean, present significant challenges. To address this, we employed open-source pretrained LLMs within the RT-Surv framework to effectively structure extensive EHR data (**Fig. 1B**).

Proprietary API-based LLMs, such as GPT-4, Gemini 1·5 Pro, and Claude 3·5 Sonnet, offer superior performance but pose significant privacy concerns due to the transmission of patient data to external corporate servers. To address these concerns, we investigated the feasibility of employing a pre-trained open-source LLM within the confines of a single institution-level resources. Specifically, we utilized Meta's LLaMA-3 model without tuning, and compared the performances of different model sizes (8B and 70B). Additionally, to evaluate the potential advantages of domain-specific models, we included a comparison with a medically fine-tuned LLM (Med-LLaMA).

We provided the LLM with expert-crafted prompts by utilizing a single-shot learning approach. The LLM then structured data from the EHRs by categorizing the patients' (1) general condition, (2) pathology classification of the primary tumor, (3) current disease extent, (4) overall disease control trend, (5) purpose of RT, (6) history of prior RT to the same site, and (7) the urgency of RT. This process of data structuring was grounded in comprehensive radiation oncology records, including referral reasons, medical history, clinical summaries, treatment plans, and the most recent imaging reports. The design of the prompts, data utilized, and classification methods are detailed in the **Supplementary Table 2**. Subsequently, we developed a predictive model incorporating both structured EHR data and LLM-structured clinical features from unstructured data, and compared its performance against a model based solely on structured data to assess the benefits of LLM-driven data structuring.

In developing the survival prediction models, the approach to handling missing data varied depending on the model utilized. For the Random Survival Forest model, missing values were not problematic due to its decision tree-based nature. However, for the Cox Proportional Hazards model and the DeepSurv model, missing values required imputation. In these models, missing data were replaced with the mean of the training dataset to facilitate effective model training and validation.

Patients were excluded from the survival prediction analysis if fewer than four variables were available from the structured electronic health record data, with the exception of LLM-structured unstructured EHR data, ensuring a fair comparison of model performance with the inclusion of LLM-structured unstructured EHR data.

The LLM employed was the LLaMA-3-70B model (Meta AI, USA), implemented using the Ollama library (Ollama, USA) in Python version 3.10 (Python Software Foundation, USA). All experiments were conducted using internal institutional computing resources, ensuring no data were exported externally. The experiments were performed on two Nvidia RTX 3090 GPUs.

**Details of single-shot learning-based structuration of unstructured EHR data**

As LLMs exhibit remarkable flexibility and adaptability, they can accurately interpret and respond to a wide variety of prompts, making them suitable for few-shot learning and even single-shot learning. This enables their application in structuring unstructured data. Therefore, we employed a single-shot learning approach by providing the LLM with specific examples of the desired tasks, classification categories, and expected outcomes, allowing the model to process new incoming data in a similar manner for structuring purposes.

To enable the LLM to predict seven variables related to patient survival in radiation oncology (general condition, primary pathology, disease extent, disease control, aim of RT, re-irradiation, and emergency RT) from unstructured data, we defined the types of unstructured data inputs for each variable. We also defined the classes into which these variables should be classified and provided an example of the expected output format. Specific prompt examples for each variable are detailed in **Supplementary Table 2**.

**Supplementary Table 1. Proportions of missing data for each electronic health record data features**

| Features | % of missing data |
|---|---|
| Age | 2.2% |
| Sex | 11.4% |
| Height | 21.7% |
| Weight | 21.1% |
| Body mass index | 25.2% |
| Systolic blood pressure | 40.9% |
| Diastolic blood pressure | 37.5% |
| Pulse rate | 36.9% |
| Body temperature | 47.6% |
| White blood cell count | 28.4% |
| Red blood cell count | 28.4% |
| Platelet count | 28.5% |
| Hemoglobin | 28.4% |
| Hematocrit | 28.1% |
| Absolute neutrophil count | 30.0% |
| Absolute lymphocyte count | 30.0% |
| Neutrophil-lymphocyte ratio | 30.1% |
| Absolute monocyte count | 30.0% |
| Absolute eosinophil count | 30.0% |
| Absolute basophil count | 30.0% |
| Calcium (plasma) | 32.3% |
| Inorganic phosphate (plasma) | 32.5% |
| Glucose (plasma) | 31.3% |
| Blood Urea Nitrogen (plasma) | 30.6% |
| Creatinine (plasma) | 30.6% |
| Uric acid (plasma) | 33.7% |
| Cholesterol (plasma) | 51.5% |
| Total protein (plasma) | 31.4% |
| Albumin (plasma) | 31.0% |
| Alkaline phosphotase (plasma) | 33.0% |
| Aspartate Aminotransferase (plasma) | 30.8% |
| Alanine Aminotransferase (plasma) | 30.8% |
| Total bilirubin (plasma) | 32.0% |
| Gamma-Glutamyl Transferase (plasma) | 87.0% |
| Sodium (plasma) | 45.9% |
| Kalium (plasma) | 45.9% |
| Chloride (plasma) | 46.5% |
| Prothrombin Time (INR, plasma) | 55.9% |
| Activated Partial Thromboplastin Tim (plasma) | 57.2% |
| Average | 34.3% |

INR, international normalized ratio.

**Supplementary Table 2. Detailed prompts used for each clinical feature**

| Clinical features | Prompts |
|---|---|
| General condition | Based on the following clinical and imaging information, please assess the general condition of the given patient. If information is not available or insufficient for accurate evaluation, classify as "9 = Not evaluable".<br>Here is the patient information and imaging study results:<br>Reason for referral: {CC};<br>Present illness: {PI};<br>Clinical note: {Note};<br>Chest X-ray: {CXR};<br>Abdomen Flat: {abdomen}.<br><br>Categorize the patient's general condition as follows:<br>- 0 = Good condition: No significant health issues, stable and healthy.<br>- 1 = Minor issues: Minor health problems, but generally stable.<br>- 2 = Moderate issues: Noticeable health problems that require medical attention.<br>- 3 = Severe issues: Serious health problems that require immediate and significant medical intervention.<br>- 9 = Not evaluable: Insufficient information to make an accurate assessment.<br><br>The following is an example of the response.<br>{{<br>"general_condition=2"<br>}}<br>Strictly follow the format of the example provided. Provide only the response itself. Do not provide any unnecessary information. |
| Primary pathology | Categorize the "pathology type of primary cancer" of the following cancer patient.<br>If information is not available or insufficient for accurate evaluation, classify as "9 = Not evaluable".<br>Here is the medical records of this patient:<br>Reason for referral: {CC};<br>Present illness: {PI};<br>Clinical note: {Note};<br>Radiotherapy plan: {Plan}.<br><br>Choose from the following options:<br>- 0 = Epithelial origin: Cancer originates from epithelial cells.<br>- 1 = Mesenchymal origin: Cancer originates from mesenchymal cells.<br>- 2 = Lymphoid and hematologic origin: Cancer originates from lymphoid or blood-forming cells.<br>- 3 = Neuroendocrine origin: Cancer originates from neuroendocrine cells.<br>- 4 = CNS origin: Cancer originates from central nervous system cells.<br>- 5 = Others: Cancer originates from other types of cells.<br>- 9 = Not evaluable: Insufficient information to make an accurate assessment.<br><br>The following is an example of the response.<br>{{<br>"pathology=0"<br>}}<br>Strictly follow the format of the example provided. Provide only the response itself. Do not provide any unnecessary information. |
| Disease extent | Based on the following clinical and imaging information, please assess the current disease (tumor) extent in the given patient. If information is not available or insufficient for accurate evaluation, classify as "9 = Not evaluable".<br>Here is the medical records and imaging study results of this patient:<br>Reason for referral: {CC};<br>Present illness: {PI};<br>Clinical note: {Note};<br>Radiotherapy plan: {Plan};<br>Brain MRI: {bMRI};<br>Chest CT: {CCT};<br>Abdomen Pelvis CT: {APCT};<br>PET-CT: {PET}.<br><br>Categorize the current disease extent as follows:<br>- 0 = No evidence of disease (NED): No detectable cancer remains.<br>- 1 = Tiny residual disease exists: Small amounts of cancer are still present.<br>- 2 = Moderate residual disease exists: Noticeable amounts of cancer are still present.<br>- 3 = Extensive & uncontrolled metastasis: Cancer has spread extensively and is not under control.<br>- 9 = Not evaluable: Insufficient information to make an accurate assessment.<br><br>The following is an example of the response.<br>{{<br>"disease_extent=2"<br>}}<br>Strictly follow the format of the example provided. Provide only the response itself. Do not provide any unnecessary information. |
| Disease control | Evaluate the "current disease control (decreasing, stable or progression)" after the previous treatment (radiotherapy, surgery or chemotherapy) of this patient. If the patient has just been initially diagnosed and has not received any cancer treatment, classify as "9 = Not evaluable".<br>Here is the medical records and imaging study results of this patient:<br>Reason for referral: {CC};<br>Present illness: {PI};<br>Clinical note: {Note};<br>Radiotherapy plan: {Plan};<br>Brain MRI: {bMRI};<br>Chest CT: {CCT};<br>Abdomen Pelvis CT: {APCT};<br>PET-CT: {PET}.<br><br>Categorize the current disease control as follows:<br>- 0 = Complete response after previous treatment: No detectable cancer remains after treatment.<br>- 1 = Partial response after previous treatment: Cancer has decreased in size but is not completely gone.<br>- 2 = Stable disease after previous treatment: Cancer has not increased or decreased significantly.<br>- 3 = Progressive disease even after previous treatment: Cancer has increased in size or spread after treatment.<br>- 9 = Not evaluable: Insufficient information to make an accurate assessment or the patient has not received any cancer treatment.<br><br>The following is an example of the response.<br>{{<br>"disease_control=2"<br>}}<br>Strictly follow the format of the example provided. Provide only the response itself. Do not provide any unnecessary information. |
| Aim of RT | Categorize the "aim of radiotherapy" of the following cancer patient. If information is not available or insufficient for accurate evaluation, classify as "9 = Not evaluable".<br>Here is the medical records of this patient:<br>Reason for referral: {CC};<br>Present illness: {PI};<br>Clinical note: {Note};<br>Radiotherapy plan: {Plan}.<br><br>Choose from the following options:<br>- 0 = Definitive or postoperative (curative): Radiotherapy is intended to cure the cancer.<br>- 1 = Salvage: Radiotherapy is intended to treat cancer that has recurred after initial treatment.<br>- 2 = Palliative: Radiotherapy is intended to relieve symptoms and improve quality of life, not to cure the cancer.<br>- 3 = Others: Radiotherapy has another purpose not covered by the above categories.<br>- 9 = Not evaluable: Insufficient information to make an accurate assessment.<br><br>The following is an example of the response.<br>{{<br>"RT_aim=0"<br>}}<br>Strictly follow the format of the example provided. Provide only the response itself. Do not provide any unnecessary information. |
| Re-irradiation | Identify whether the patient has previously received radiation therapy in the area currently being considered for radiation treatment. If information is not available or insufficient for accurate evaluation, classify as "9 = Not evaluable".<br>Here is the medical records of this patient:<br>Reason for referral: {CC};<br>Present illness: {PI};<br>Clinical note: {Note}.<br><br>Consider only whether the patient has previously received radiation therapy to the same area, excluding the current treatment being considered. If radiation therapy has not been previously administered to the same area that currently requires treatment, it is not considered re-RT.<br><br>Choose from the following options:<br>- 0 = No: The patient has not received previous radiation therapy to the same area.<br>- 1 = Yes: The patient has received previous radiation therapy to the same area.<br>- 9 = Not evaluable: Insufficient information to make an accurate assessment.<br><br>The following is an example of the response.<br>{{<br>"re_RT=0"<br>}}<br>Strictly follow the format of the example provided. Provide only the response itself. Do not provide any unnecessary information. |
| Emergency | Assess the severity of the patient's symptoms and their current general condition. Consider whether the patient's condition falls under indications for emergency radiation therapy, such as cord compression, SVC syndrome, acute hemorrhage, or brain metastasis with severe symptoms. Based on this comprehensive assessment, determine the urgency of the patient's need for radiation therapy. If information is not available or insufficient for accurate evaluation, classify as "9 = Not evaluable".<br>Here is the medical records of this patient:<br>Reason for referral: {CC};<br>Present illness: {PI};<br>Clinical note: {Note};<br>Radiotherapy plan: {Plan}.<br><br>Categorize the urgency of the need for radiation therapy as follows:<br>- 0 = Not emergent at all. More like elective treatment: The patient does not require immediate radiation therapy.<br>- 1 = Slightly emergent, with some symptoms: The patient has symptoms that may benefit from timely radiation therapy.<br>- 2 = Moderately emergent, treatment needed soon. Moderately significant symptoms: The patient has significant symptoms requiring prompt radiation therapy.<br>- 3 = Emergent treatment needed immediately: The patient has severe symptoms requiring immediate radiation therapy.<br>- 9 = Not evaluable: Insufficient information to make an accurate assessment.<br><br>The following is an example of the response.<br>{{<br>"emergency=1"<br>}}<br>Strictly follow the format of the example provided. Provide only the response itself. Do not provide any unnecessary information. |

**Supplementary Table 3. Patient characteristics from structured electronic health record data according to 30-day mortality (DM) after radiotherapy (RT)**

| Features | Patients without 30-DM | | Patients with 30-DM | | p-value |
|---|---|---|---|---|---|
| | n or mean | % or STD | n or mean | % or STD | |
| Age (years) | 59.5 | | 61.8 | | <0.001 |
| Sex | | | | | <0.001 |
| Female | 15387 | 46.4% | 361 | 32.2% | |
| Male | 14256 | 43.0% | 665 | 59.3% | |
| Not available | 3511 | 10.6% | 96 | 8.6% | |
| Height (cm) | 162.0 | ± 10.7 | 162.7 | ± 11.5 | 0.033 |
| Weight (kg) | 61.5 | ± 12.0 | 58.7 | ± 11.4 | <0.001 |
| Body mass index (kg/m²) | 23.4 | ± 5.2 | 22.0 | ± 4.8 | <0.001 |
| Systolic blood pressure (mmHg) | 124.1 | ± 15.9 | 125.5 | ± 16.8 | 0.013 |
| Diastolic blood pressure (mmHg) | 75.8 | ± 11.0 | 78.3 | ± 11.9 | <0.001 |
| Pulse rate (beats per minute) | 78.0 | ± 17.3 | 86.8 | ± 24.7 | <0.001 |
| Body temperature (°C) | 36.8 | ± 0.4 | 36.9 | ± 0.4 | <0.001 |
| White blood cell count (×10³/μL) | 7.2 | ± 4.3 | 10.1 | ± 7.0 | <0.001 |
| Red blood cell count (×10⁶/μL) | 3.9 | ± 0.7 | 3.5 | ± 0.7 | <0.001 |
| Platelet count (×10³/μL) | 250.5 | ± 109.1 | 230.9 | ± 133.7 | <0.001 |
| Hemoglobin (g/dL) | 11.8 | ± 2.0 | 10.6 | ± 2.0 | <0.001 |
| Hematocrit (%) | 35.4 | ± 5.8 | 31.6 | ± 6.1 | <0.001 |
| Absolute neutrophil count (×10³/μL) | 4.9 | ± 3.6 | 8.1 | ± 6.7 | <0.001 |
| Absolute lymphocyte count (×10³/μL) | 1.6 | ± 1.3 | 1.1 | ± 0.7 | <0.001 |
| Neutrophil-lymphocyte ratio | 4.2 | ± 5.4 | 10.5 | ± 15.7 | <0.001 |
| Absolute monocyte count (×10³/μL) | 0.6 | ± 0.7 | 0.7 | ± 0.5 | <0.001 |
| Absolute eosinophil count (×10³/μL) | 0.2 | ± 0.2 | 0.1 | ± 0.4 | 0.538 |
| Absolute basophil count (×10³/μL) | 0.0 | ± 0.0 | 0.0 | ± 0.0 | 0.613 |
| Calcium (mg/dL) | 9.1 | ± 0.6 | 8.9 | ± 0.9 | <0.001 |
| Inorganic phosphate (mg/dL) | 3.6 | ± 0.7 | 3.3 | ± 0.9 | <0.001 |
| Glucose (mg/dL) | 118.5 | ± 44.0 | 130.2 | ± 48.2 | <0.001 |
| Blood Urea Nitrogen (mg/dL) | 15.5 | ± 6.9 | 18.6 | ± 12.1 | <0.001 |
| Creatinine (mg/dL) | 0.8 | ± 0.5 | 0.8 | ± 0.6 | 0.800 |
| Uric acid (mg/dL) | 4.4 | ± 1.6 | 4.0 | ± 2.1 | <0.001 |
| Cholesterol (mg/dL) | 168.4 | ± 45.0 | 153.4 | ± 51.1 | <0.001 |
| Total protein (g/dL) | 6.7 | ± 0.8 | 6.2 | ± 0.9 | <0.001 |
| Albumin (g/dL) | 4.0 | ± 0.6 | 3.3 | ± 0.7 | <0.001 |
| Alkaline phosphatase (U/L) | 109.4 | ± 133.6 | 216.1 | ± 262.5 | <0.001 |
| Aspartate Aminotransferase (U/L) | 30.5 | ± 42.4 | 57.9 | ± 92.0 | <0.001 |
| Alanine Aminotransferase (U/L) | 27.0 | ± 35.3 | 32.0 | ± 48.1 | <0.001 |
| Total bilirubin (mg/dL) | 0.6 | ± 0.9 | 1.3 | ± 2.7 | <0.001 |
| Gamma-Glutamyl Transferase (U/L) | 135.1 | ± 197.9 | 232.9 | ± 272.7 | <0.001 |
| Sodium (mEq/L) | 138.6 | ± 3.5 | 135.2 | ± 4.8 | <0.001 |
| Potassium (mEq/L) | 4.3 | ± 0.5 | 4.2 | ± 0.6 | |
| Chloride (mEq/L) | 102.4 | ± 4.1 | 99.6 | ± 5.3 | <0.001 |
| Prothrombin Time (INR) | 1.0 | ± 0.2 | 1.2 | ± 0.3 | <0.001 |
| aPTT (seconds) | 30.9 | ± 5.8 | 30.7 | ± 6.9 | 0.426 |

STD, standard deviation; INR, international normalized ratio; Activated Partial Thromboplastin Time.

**Supplementary Table 4. Example of input data, corresponding prompt, and responses from various large language models**

| Chest CT | Abdominopelvic CT | Clinical note | Prompt | LLaMA-3-70B | LLaMA-3-8B | Med-LLaMA-2-7B |
|---|---|---|---|---|---|---|
| Slightly decreased irregular mass-like wall thickening in the lower esophagus.<br>Slightly increased small nodules in the left upper lobe (lingular) and left lower lobe (posterior basal), R/O metastasis.<br>Underlying emphysema and COPD.<br>No significant change in metastatic lymph nodes (supraclavicular, bilateral paratracheal, subcarinal, paraesophageal, and around EG junction).<br>Impression: Overall stable disease. | Multiple subcentimeter low-attenuating lesions in the liver, possible metastasis.<br>Decreased sizes of a few liver metastases.<br>Decreased sizes of metastatic LNs in the lower paraesophageal and left gastric areas.<br>Atherosclerosis along the aorta and its branches.<br>No change in bilateral renal cysts.<br>Pelvic ascites.<br>Refer to Chest CT for the included lower thorax.<br>Impression:<br><br>Overall decreased metastases in the liver and lower paraesophageal/left gastric LNs.<br>Multiple subcentimeter low-attenuating lesions in the liver, possible metastasis. | Reason for Referral: Consolidation RT<br><br>#1. Esophagus, UI 30-42cm, Small cell neuroendocrine ca, cT+N2/3M1 (2024.03)<br>T: Irregular mass in lower esophagus, suggestive of esophageal cancer. Increased FDG uptake in distal esophagus, consistent with primary cancer.<br>N: Enlarged metastatic lymph nodes in supraclavicular, bilateral paratracheal, subcarinal, paraesophageal areas, and around EG junction. Metastatic LNs at LGA and Rt. paracardial area.<br>M: 2cm rim-enhancing lesion in S4a liver, likely metastasis.<br>s/p Esophagus, EGD biopsy<br><br>Poorly differentiated malignant tumor, suggestive of small cell neuroendocrine ca<br><br>On E.P.S/Cisplatin<br><br>Multidiciplinary clinic:<br>Significant tumor reduction post-chemotherapy<br>Residual tumor to receive consolidation RT | Categorize the current disease extent as follows:<br>- 0 = No evidence of disease (NED): No detectable cancer remains.<br>- 1 = Tiny residual disease exists: Small amounts of cancer are still present.<br>- 2 = Moderate residual disease exists: Noticeable amounts of cancer are still present.<br>- 3 = Extensive & uncontrolled metastasis: Cancer has spread extensively and is not under control.<br>- 9 = Not evaluable: Insufficient information to make an accurate assessment.<br><br>The following is an example of the response.<br>{{<br>"disease_extent=2"<br>}}<br>Strictly follow the format of the example provided.<br>Provide only the response itself. Do not provide any unnecessary information. | {{<br>"disease_extent=1"<br>}} | {{<br>"disease_extent=2"<br>}} | Giventhatthecurrentdiseaseextentisstable,thepatientshouldbefollowedupregularlywithCTorotherimagingstudiesasclinicallyindicated.Dependingonthepatient'sconditionandresponsetotreatment,furtherRTmaybeconsidered.However,itisessentialtonotethatthisinformationissubjecttochangedependingonfuturestudies. |

CT, computed tomography; B, billion parameters.

**Supplementary Table 5. Patient characteristics from structured electronic health record (EHR) data according to 30-day mortality (DM) after radiotherapy (RT)**

| Variables | Patients without 30-DM | | Patients with 30-DM | | p-value |
|---|---|---|---|---|---|
| | n | % | n | % | |
| General condition | | | | | <0.001 |
| Good | 48 | 0.1% | 0 | 0.0% | |
| Minor issue | 5746 | 17.3% | 2 | 0.2% | |
| Major issue | 9380 | 28.3% | 17 | 1.5% | |
| Severe issue | 17979 | 54.2% | 1103 | 98.3% | |
| Not evaluable | 1 | 0.0% | 0 | 0.0% | |
| Primary pathology | | | | | <0.001 |
| Epithelial origin | 27603 | 83.3% | 971 | 86.5% | |
| Mesenchymal origin | 903 | 2.7% | 26 | 2.3% | |
| Lymphoid and hematologic origin | 1284 | 3.9% | 42 | 3.7% | |
| Neuroendocrine origin | 1022 | 3.1% | 66 | 5.9% | |
| CNS origin | 2227 | 6.7% | 11 | 1.0% | |
| Others | 51 | 0.2% | 4 | 0.4% | |
| Not evaluable | 64 | 0.2% | 2 | 0.2% | |
| Disease extent | | | | | <0.001 |
| No evidence of disease | 3668 | 11.1% | 4 | 0.4% | |
| Tiny amount residual disease | 12744 | 38.4% | 28 | 2.5% | |
| Moderate amount of residual disease | 7766 | 23.4% | 85 | 7.6% | |
| Extensive metastasis | 8477 | 25.6% | 1000 | 89.1% | |
| Not evaluable | 499 | 1.5% | 5 | 0.4% | |
| Disease control | | | | | <0.001 |
| Complete response | 2388 | 7.2% | 1 | 0.1% | |
| Partial response | 8120 | 24.5% | 54 | 4.8% | |
| Stable disease | 4622 | 13.9% | 53 | 4.7% | |
| Progressive disease | 10490 | 31.6% | 909 | 81.0% | |
| Not evaluable | 7534 | 22.7% | 105 | 9.4% | |
| Aim of RT | | | | | <0.001 |
| Curative (definitive/postoperative) | 18939 | 57.1% | 64 | 5.7% | |
| Salvage | 5675 | 17.1% | 61 | 5.4% | |
| Palliative | 8287 | 25.0% | 995 | 88.7% | |
| Others | 246 | 0.7% | 2 | 0.2% | |
| Not evaluable | 7 | 0.0% | 0 | 0.0% | |
| Re-irradiation | | | | | <0.001 |
| No | 26571 | 80.1% | 743 | 66.2% | |
| Yes | 6300 | 19.0% | 352 | 31.4% | |
| Not evaluable | 283 | 0.9% | 27 | 2.4% | |
| Emergency | | | | | <0.001 |
| Not emergent at all | 11676 | 35.2% | 10 | 0.9% | |
| Slightly emergent | 3964 | 12.0% | 17 | 1.5% | |
| Moderately emergent | 16378 | 49.4% | 827 | 73.7% | |
| Indicative for emergency RT | 1136 | 3.4% | 268 | 23.9% | |

**Supplementary Fig. 1. Flow chart of patient selection and data split**

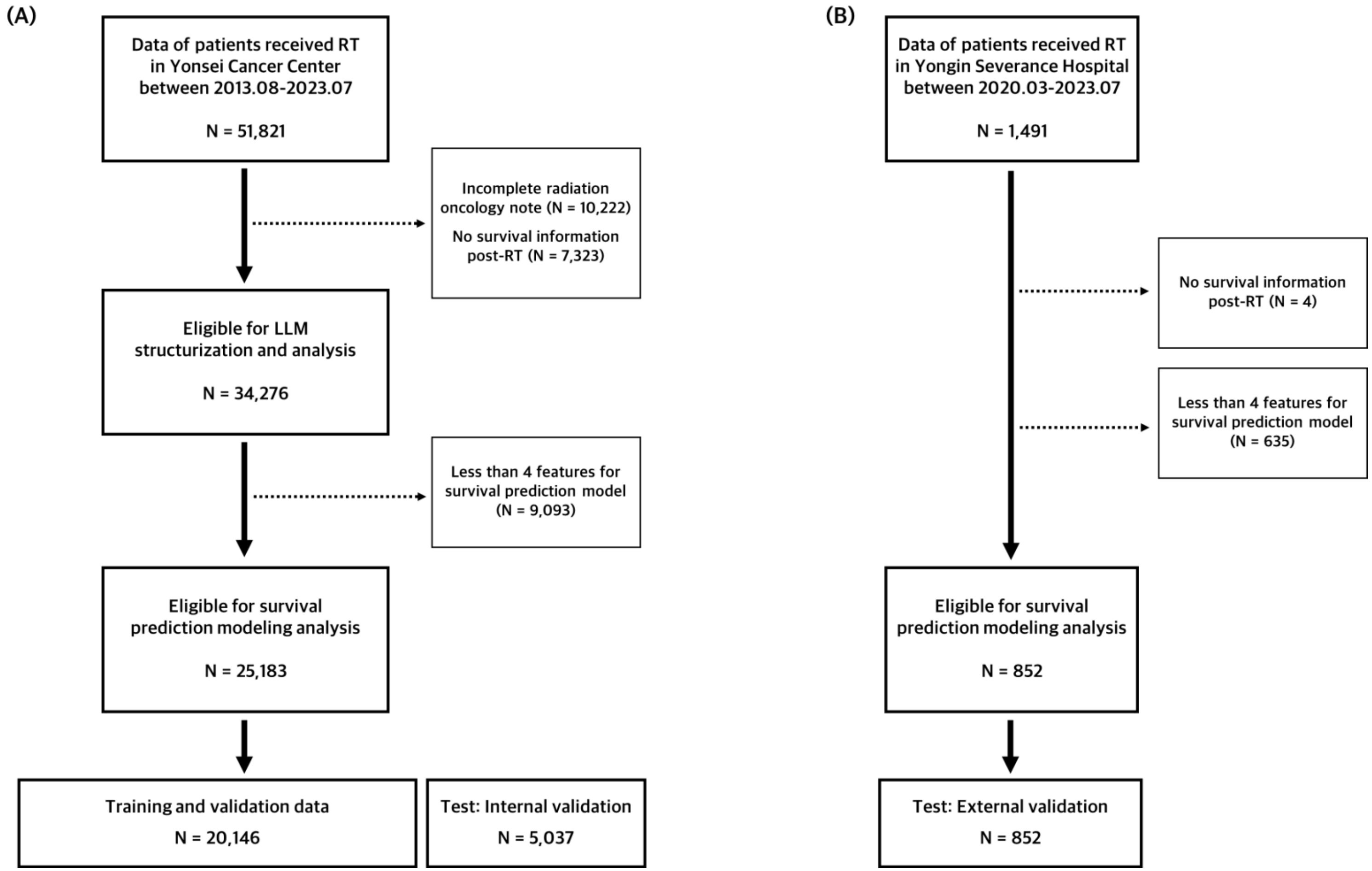

Flow chart of patient selection and data split. Flow chart for (A) training and internal validation data and (B) external validation data.

**Supplementary Fig. 2. Data collection timepoints of electronic health record (EHR) data**

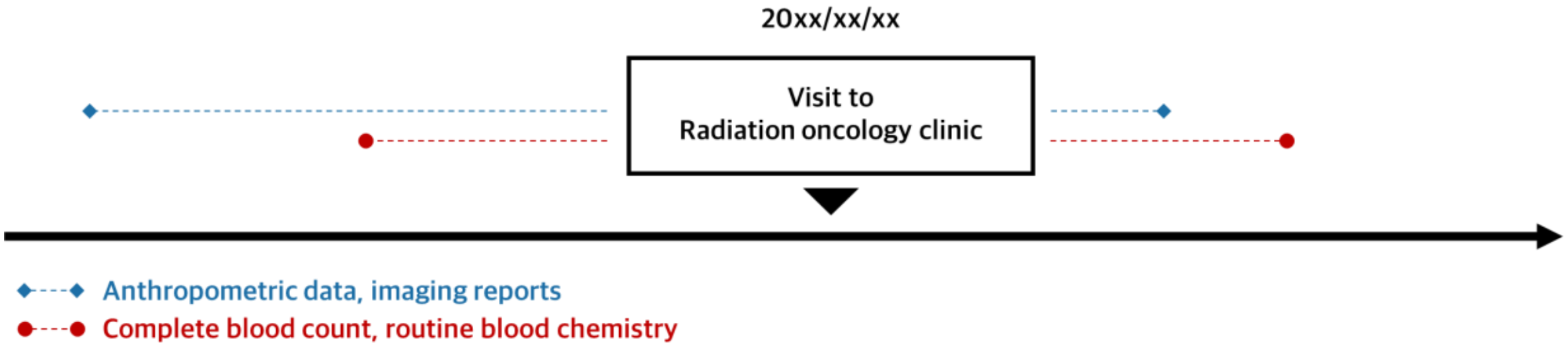

Data collection timepoints for each EHR feature relative to the patient's visit to the radiation oncology clinic for radiotherapy referral. Anthropometric data and imaging reports were collected from 28 days before to 7 days after the visit. Complete blood count and routine blood chemistry were collected from 14 days before to 14 days after the visit. The data closest to the clinic visit date were used for analysis.

**Supplementary Fig. 3. Comparison of large language model-structured electronic health record (EHR) data according to 30-day mortality (30-DM) after radiotherapy (RT)**

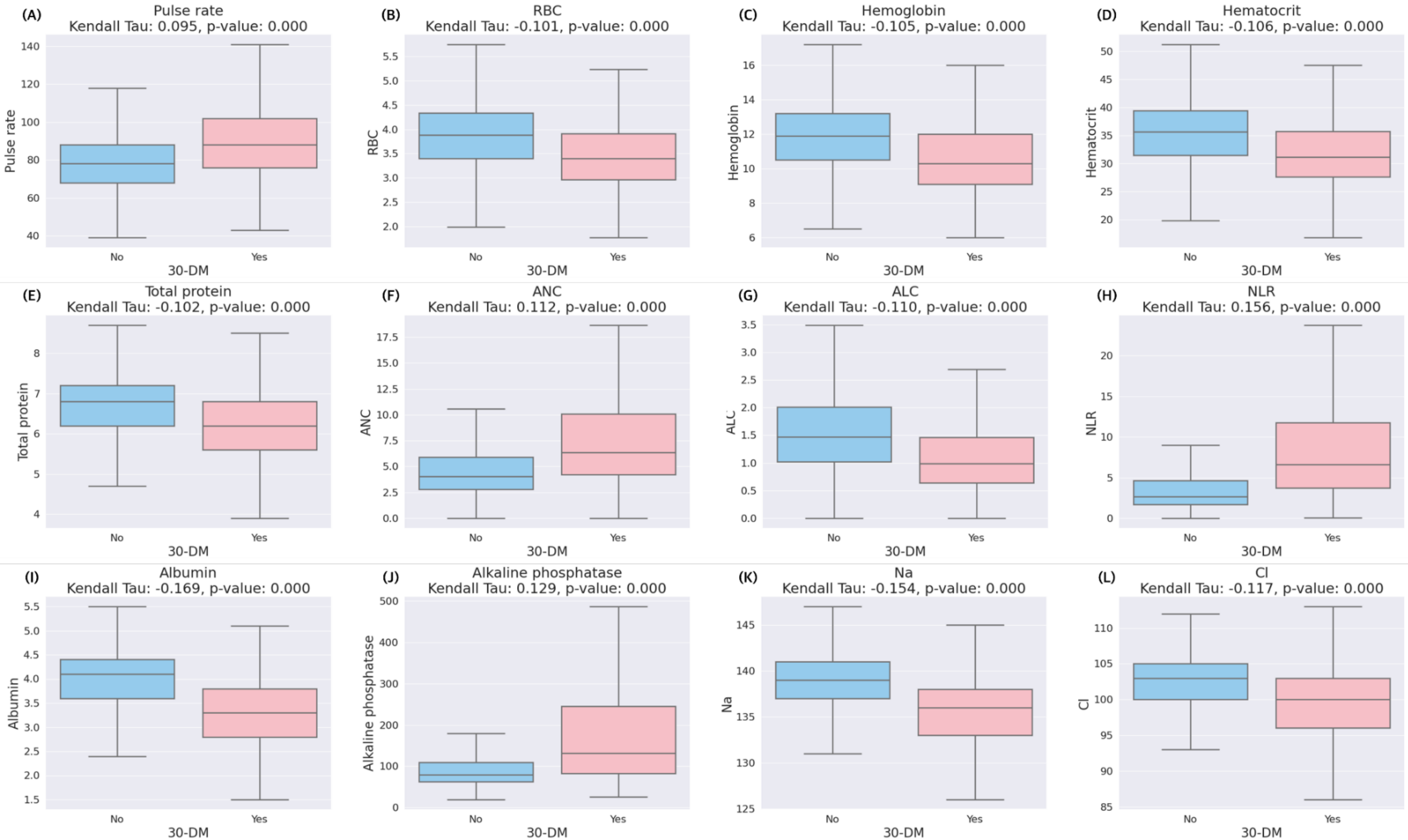

Comparison of various structured EHR according to 30-DM occurrence after RT. The boxplots show the distribution of each data for patients with and without 30-DM. Data include (A) pulse rate, (B) red blood cell count (RBC), (C) hemoglobin, (D) hematocrit, (E) total protein, (F) absolute neutrophil count (ANC), (G) absolute lymphocyte count (ALC), (H) neutrophil-to-lymphocyte ratio (NLR), (I) albumin, (J) alkaline phosphatase, (K) sodium (Na), and (L) chloride (Cl). Kendall's Tau and p-values indicate the strength and significance of the association between each data and 30-DM.

**Supplementary Fig. 4. Performance comparison of survival prediction models in internal and external validation**

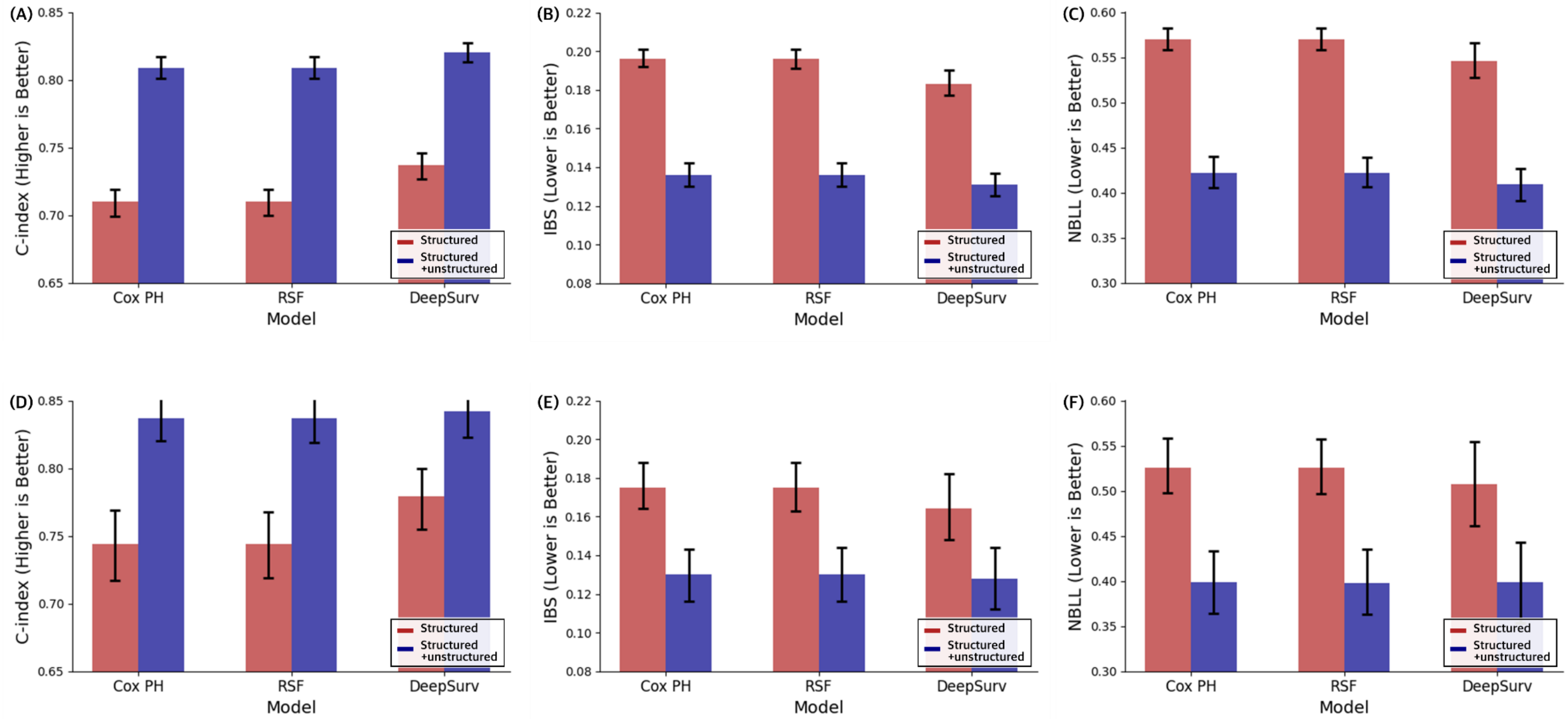

The survival prediction performance was evaluated for models using structured data only (red) versus models integrating both structured and unstructured data (blue). Panels (A-C) represent the results from internal validation: (A) C-index (higher is better), (B) Integrated Brier Score (IBS, lower is better), and (C) Negative Binomial Log-Likelihood (NBLL, lower is better). Panels (D-F) show the corresponding results from external validation: (D) C-index, (E) IBS, and (F) NBLL. The assessed models included Cox proportional hazards (Cox PH), random survival forests (RSF), and DeepSurv. The error bars denote 95% confidence intervals for each metric.